\documentclass[runningheads]{llncs}
\usepackage{booktabs}
\usepackage{wrapfig}
\usepackage[T1]{fontenc}
\usepackage{graphicx}
\usepackage{hyperref}
\usepackage{enumitem}
\usepackage{xcolor}
\usepackage{amsfonts,amssymb}

\definecolor{chromeyellow}{rgb}{1.0, 0.65, 0.0}

\begin{document}

\title{Literal-Aware Knowledge Graph Embedding for Welding Quality Monitoring: A Bosch Case}
\titlerunning{Literal-Aware KGE for Welding Quality Monitoring}

\author{Zhipeng Tan\inst{1,2,}\thanks{Contact email addresses of corresponding authors: \email{zhipeng.tan@rwth-aachen.com}, \email{baifanz@ifi.uio.no}, \email{evgeny.kharlamov@de.bosch.com}. Code and data are available  under \url{https://github.com/boschresearch/KGE-Welding}}  \and
Baifan Zhou\inst{3,4,*}\and
Zhuoxun Zheng\inst{1,3} \and 
Ognjen Savkovic\inst{5} \and \\
Ziqi Huang\inst{2} \and
Irlan-Grangel Gonzalez\inst{1} \and
Ahmet Soylu\inst{4} \and \\
Evgeny Kharlamov\inst{1,3,*}
}
\authorrunning{Tan et al.}
\institute{
Bosch Center for AI, Germany  \and
RWTH Aachen University, Germany  \and
Department of Informatics, University of Oslo, Norway \and
Department of Computer Science, Oslo Metropolitan University, Norway  \and
Department of Computer Science, Free University of Bozen-Bolzano, Italy
}

\maketitle              
\begin{abstract}
Recently there has been a series of studies in knowledge graph embedding (KGE), which attempts to learn the embeddings of the entities and relations as numerical vectors and mathematical mappings via machine learning (ML). 
However, there has been  limited research that applies KGE for industrial problems in manufacturing. This paper investigates whether and to what extent KGE can be used for an important problem: quality monitoring for welding in manufacturing industry, which is an impactful process accounting for production of millions of cars annually. 
The work is in line with Bosch research of data-driven solutions that intends to replace the traditional way of destroying cars, which is extremely costly and produces waste. 
The paper tackles two very challenging questions simultaneously: how large the welding spot diameter is; and to which car body the welded spot belongs to.
The problem setting is difficult for traditional ML because there exist a high number of car bodies that should be assigned as class labels. We formulate the problem as link prediction, and experimented popular KGE methods on real industry data,  with consideration of literals.
Our results reveal both limitations and promising aspects of adapted KGE methods.
\looseness=-1

\end{abstract}

\section{Introduction}
\label{sec:intro}

\begin{figure*}[t]
\centering
\includegraphics[width=.8\textwidth]{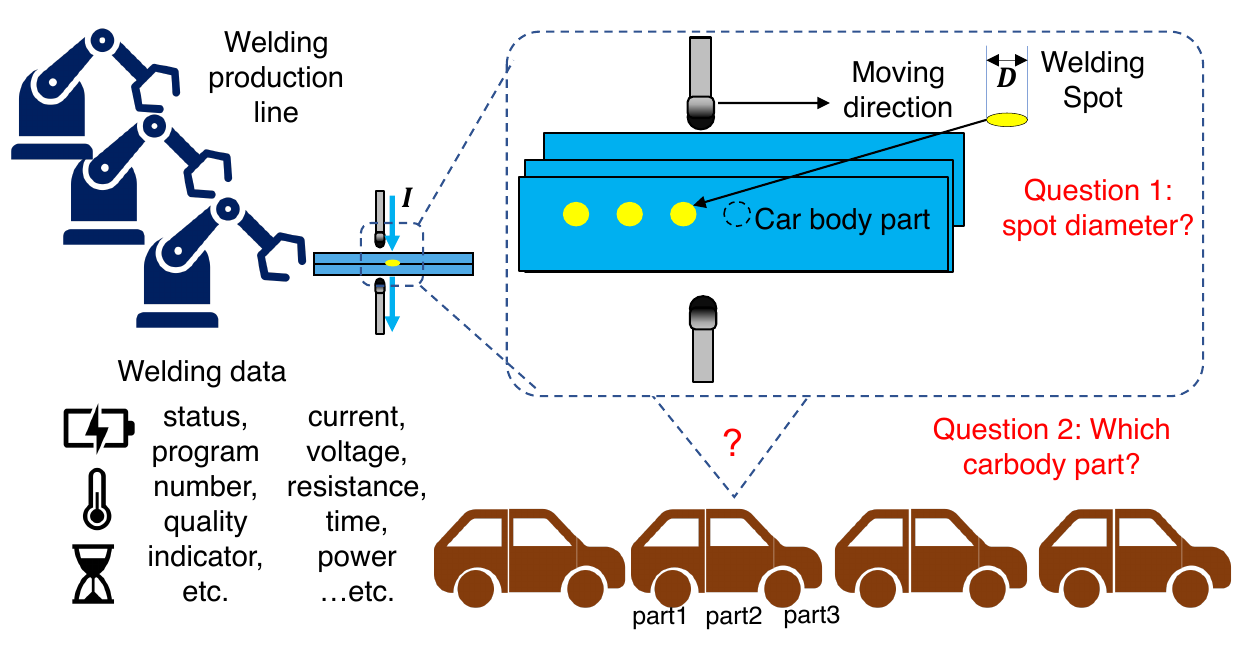}
\vspace{-4mm}
\caption{Two core questions in welding quality monitoring: \textit{Question 1} (Q1), how large is the spot diameter? \textit{Question 2} (Q2), which car body part does this spot diameter belongs to?}
\label{fig:weldingmonitor}
\vspace{-2ex}
\end{figure*}

\noindent \textbf{Background and Challenge.}
In automotive industry, automated welding is essential for manufacturing high-quality car bodies, accounting for over millions of car production annually. Welding is a data-intensive process. Considering the production lines in Fig.~\ref{fig:weldingmonitor} with 10-20 welding machines in each line, each welding machine produces one spot in several second or minutes, and a car body can have up to 6000 spots~\cite{zhou2022machine}. For each spot, several hundreds of features are generated, including welding status, quality indicators, and sensor measurements, where the sensors measure important physical properties every millisecond, such as current, resistance, power. 

This large amount of data increases the demand of data-driven solutions~\cite{2023zhousemcloud,huang2023hybrid}, which aims  to reduce and eventually replace  conventional destructive methods. In the case of the latter, only a small sample of welded car bodies can be measured, because the sample needs to be destroyed, making the methods to be extremely expensive and also producing waste.
Two core questions need to be answer here as shown in Fig.~\ref{fig:weldingmonitor}. Q1 is important because the spot diameter is the key quality indicator for judging welding quality. It must be above a certain threshold, because a too small diameter means insufficient connection and can cause severe consequences (e.g., car user safety).
The diameter should also not be too large, because this means energy inefficiency and can cause quality deficiency of the surrounding spots (by e.g. short-circuit effect).
Q2 is important because it is essential to know the percentage of good spots for each car body, and this percentage must be higher than certain thresholds according to quality standards.
Bosch is doing research to develop data-driven solutions~\cite{zhou2022device} and semantic technologies~\cite{zheng2022towards,zhou2022towards,rincon2023addressing} for answering the two questions, whereby both classic machine learning (ML)~\cite{zhou2022machine} and the recent methods of knowledge graph embedding (KGE) are under consideration.

\medskip
\noindent
\textbf{Knowledge Graph Embedding.}
There has been a series of research on KGE methods recently~\cite{zheng2020dgl,ali2021pykeen,li2021learning}. 
In essence, KGE attempts to represents nodes and edges in KGs as vectors/matrices or mathematical mappings.
Mainstream works include translational models~\cite{transe}, bilinear models~\cite{yang2015embedding}, graph neural networks~\cite{GNN}, etc. They have studied KGE for downstream tasks such as link prediction~\cite{wang2021survey}, entity classification~\cite{yan2022survey}, and entity alignment~\cite{sun2018bootstrapping}.
 KGE for industrial applications is a relatively new trend. 
We observe that there has been limited investigation done especially in the area of  KGE for manufacturing industries (to our best knowledge). 
One recent work applied KGE on ecotoxicological effect prediction~\cite{iswc2019inuse}, where the KGE models are applied to enhance the MLP model  for predicting the effects of chemical compounds on specific species. Other works applied KGE in text processing~\cite{Santini2022}, where the user names and name abbreviations are mapped to the same author based on their publications. 
Inspired by these works, we consider it an interesting and important research question to study whether and to what extent KGE can be applied for manufacturing industry.\looseness=-1

\medskip
\noindent \textbf{Our Approach.}
To this end, we investigate KGE for answering the two questions in the automotive industry with Bosch data, and compare with a representative classic ML method. In this work,
(1) we first give a detailed introduction of the welding quality monitoring use case and welding data (Sect.~\ref{sec:usecase});
(2) after that, we formulate regression and classification problems as link prediction problems (Sect.~\ref{sec:method});
(3) and construct KGs from tabular data, during which we pay special attention to handling literals, and discretise the literals in intervals and create entities on them;
(4) we compare mainstream KGE methods such as TransE, RotatE, AttH with multilayer perceptron (MLP) (Sect.~\ref{sec:eva}). In addition, we conduct an ablation study to investigate whether the literals are important. Furthermore, we compare a variant KGE method proposed in a recent application paper~\cite{iswc2019inuse} to see if this  method is applicable;
(5) we introduce adapted performance metrics to increase the applicability of KGE 
to our industrial problems and give recommendations for further adoption in these settings (Sect.~\ref{sec:uptake}).

\section{Use Case}
\label{sec:usecase}

\noindent \textbf{Welding process.}
We refer to automated welding as a family of manufacturing processes where multiple metal parts are melted and then connected together.
An example process and production line is illustrate in Fig~\ref{fig:weldingmonitor}, during which, the welding robots control welding electrodes move along the car body parts. A high current (several kilo Ampere) passes through the electrodes and car bodies, generates heat on the metal, melting the metal to produce welding nuggets for connecting the car bodies.
 Welding is heavily applied in automotive industry, accounting for over millions of carbody production annually, where which carbody has upto 6000 spots.
Monitoring welding quality has been a key problem for industrial manufacturing, due to the requirement of accuracy and efficiency at the same time. \textit{Traditional quality monitoring} applies destructive testing, where the test carbody is intentionally destroyed to evaluate its properties and performance, which is timely and financially expensive, and produces waste, considering the large amount of welding spot. Furthermore, this only covers part of the welding quality, because the destroyed car bodies can not be used as product. 
Thus, for large scale production, reliable, highly automatic and efficient quality monitoring method with data-driven model would be preferred to replace the traditional destructive testing. 

\begin{table}[t]
    \centering
    \caption{Example of welding data, including three tables: (a) \textit{main protocol} with ids, status, etc.; (b) \textit{welding meta setting} with the carbody component information such as type, material; and (c) \textit{sensor measurements}, which are the physical properties measured per milliseconds. Note that the carbody information does not exist in the welding table data.}
    \label{table:weldingdata}
    \vspace{-2mm}
    \resizebox{\textwidth}{!}{
    \includegraphics{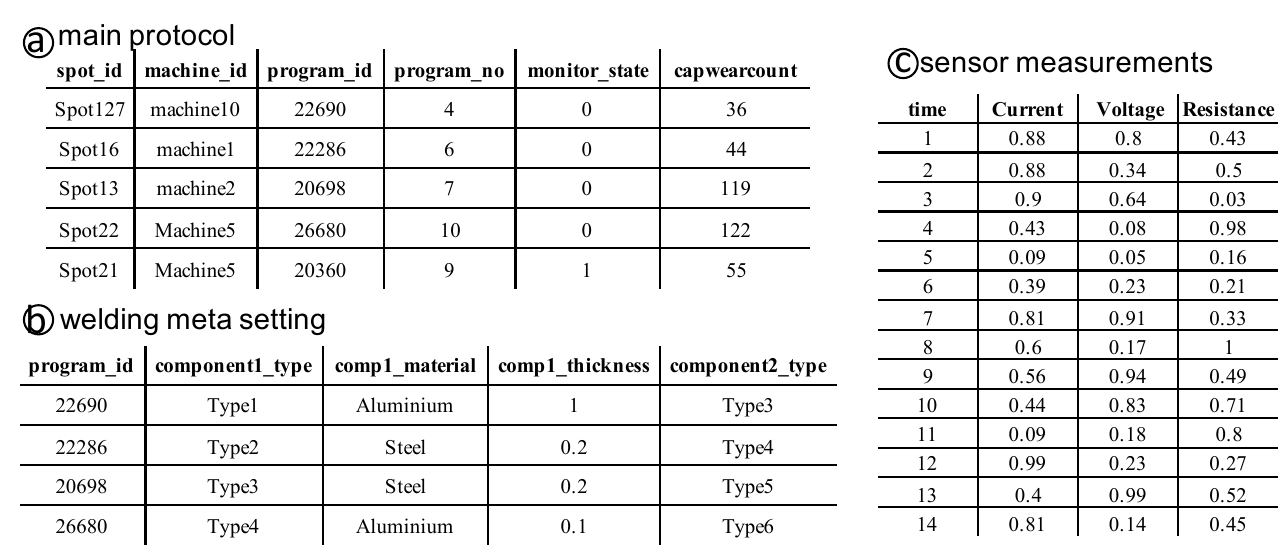}
    }
   \vspace{-10mm}
\end{table}

\medskip
\noindent \textbf{Welding Data.} 
Welding data include various information gathered from the welding process (Fig.~\ref{fig:weldingmonitor}).
We exemplify welding data with the three tables in Table~\ref{table:weldingdata}. 
Welding machines are installed with different sensors which can detect the parameter values of all the machines, these data will be collected and stored in various sources, including the the welding protocol, the welding setting database and the sensor measurements. Table.~\ref{table:weldingdata} shows an example of the recorded welding data with selected columns. The welding protocol involves all the parameters used for the machines when conducting the welding process, such as the welding machines, the welding program, the welding state etc. This welding setting database (metadata) contain information about the materials being welded, such as the type, materials and thickness of welded sheets. The sensor measurements contain all the numeric literals that are measured during the welding process over different time span, including the current, the voltage, PWM and the resistance. The welding data we are utilising are the combination of data from welding protocol, welding setting database, and the sensor measurements.
The details of the columns of Table~\ref{table:weldingdata} are given as follows: 

\begin{itemize}[topsep=3pt,parsep=0pt,partopsep=0pt,itemsep=0pt,leftmargin=*]
    \item \textit{Welding Machine} records the machines that perform the welding operations. 
    \item \textit{Welding Program} is the program installed in the welding machine used for different welding operation. Literals are the data measured by sensors in the welding process, including current, voltage, resistance, power, and other important sensor measurements. 
    \item \textit{Component Type} is the components of the welding spot. A welding spot will connect a few sheets (components), where each sheet has an impact on the resulting welding spot diameter. The three different components are also closely related to the carbody.    
\end{itemize}

Welding data is important for ensuring that the welding process is performed correctly and that the resulting weld is strong and durable. In our work, welding data will be used for quality control and quality monitoring purposes, such as monitoring the location of the welding spots and the diameters. 

\medskip \noindent \textbf{Data Anonymisation.} 
Data anonymisation and data simulation are important approaches to keep the privacy of the data.
Because Bosch has strict regulations that protects the company privacy, and  the production data contain numeric values subject potential leak of confidential information, the production data are not directly disclosed in the open data set for KGE.
We anonymise the production data from a factory in Germany and simulate part of the data based on domain knowledge, so that the data capture the statistics of the real data but do not disclose any potentially confidential information.  We conducted the anonymisation and the discretisation on the numeric values, which follows the idea of literal embedding of previous works~\cite{10.1007/978-3-030-30793-6_20,Wang2022AugmentingKG}. We provide the anonymised dataset in the open source Github repository, aiming at improving the reproducibility of the work, and potential reuse for investigation of KG embedding.

\section{Preliminaries}
\label{sec:prelim}

\textbf{Knowledge Graph (KG)} 
represented as $G = (\mathcal{E}, 
\mathcal{R},
\mathcal{L})$ is a graph-structured data model, where $\mathcal{E}$ is a set of entities, $\mathcal{L}$ is a set of literals included in the knowledge graph, and $\mathcal{R}$ is a set of binary relations, which can be further divided into two groups $\mathcal{R} = \{\mathcal{P}_{o}, \mathcal{P}_{d}\}$, where $\mathcal{P}_{o}$ denotes the relations between entities ($\mathcal{P}_{o} \subseteq \mathcal{E}\times\mathcal{E}$,  known as object properties), while $\mathcal{P}_{d}$ denotes the relations between entities and literals ($\mathcal{P}_{d} \subseteq \mathcal{E}\times\mathcal{L}$, known as datatype properties).

\medskip
\noindent \textbf{Knowledge Graph Embedding (KGE)}
in a common setting~\cite{chami2020low},
seeks to find a function (also model) that represents entities ($e \in \mathcal{E}$) as vectors ($v_{e} \in \mathcal{U}^{e}$)
and relations ($r \in \mathcal{R}$) as mathematical mappings  ($r_{r} \in \mathcal{U}^{r}$),  with a given set of triples $(h,r,t)\in \mathcal{T} \subseteq \mathcal{E} \times \mathcal{R} \times \mathcal{E}$, where the $\mathcal{U}^{e}$ and $\mathcal{U}^{r}$ are some choices of embedding spaces for entities and relations, respectively.
Commonly a KGE model is trained with ML by solving the problem of link prediction: $(h,r,?)$, namely given the \textit{query} of the head entity $h \in \mathcal{E}$ and the relation $r \in \mathcal{R}$, to find the most probable tail entity $t$ (for simplicity, we denote the query in both two directions as $(h,r, ?)$). 
Thus, a KGE model needs to find a scoring function $s:\mathcal{E} \times \mathcal{R} \times \mathcal{E} \rightarrow \mathbb{R}$, which measures the plausibility of a triple $(h,r,t)$.
In literal-aware KGE, the triples are in the form of
$(h,r,t)\in \mathcal{T} \subseteq 
\mathcal{E} \times \mathcal{R} \times \{\mathcal{E},\mathcal{L}\}$,
and the relations have two groups ($r \in \mathcal{R} = \{\mathcal{P}_{o},\mathcal{P}_{d}\}$). Both the literals and the relations need to be handled properly.
We list some popular KGE models below.

\medskip \noindent
\textit{TransE} ~\cite{transe} represents entities as vectors and relations as translation operations between these vectors. 
Specifically, 
given a query $(h, r, ?)$ in a KG, TransE predicts the tail entity as $f(h,t) = v_{h} + v_{r}$.
TransE then minimizes the Euclidean distance between the predicted and the true entity representation while maximizing the distance between the predicted and the false entity representation: $Dist(v'_{t}, v_{t}) \rightarrow 0$ for true tail entity, while $Dist(v'_{t}, v_{e}) \rightarrow MAX$ for other entities except the true tail.

\medskip \noindent
\textit{DistMult}~\cite{yang2015embedding} 
models the interactions between entities and relations as dot products in a low-dimensional space. Specifically, 
The score function is calculated in the matrix multiplication 
$g_r(h,t) = v^{\mathsf {T}}_h \cdot Mr\cdot v_t$
, here $^{\mathsf {T}}$ denotes the transpose, while 
DistMult maximizes the score for true triples while minimizing the scores for negative triples, where the score 

\medskip \noindent
\textit{RotatE}~\cite{sunrotate} is similar to TransE, but models the relations as rotation vectors. 
Specifically, 
tail entities are predicted from head entities and relations through $f(h,t) = v_{h} \circ v_{r}$, where $\circ$ denotes rotation in complex number space. The distance function is defined as cosine distance $Dist(h,t) = cos(v'_{t}, v_{t})$.

\medskip \noindent
\textit{AttH}~\cite{chami2020low} models the relations as reflections and rotations in the hyperbolic space as well as weights in the attention mechanism which combines the two hyperbolic transformations. 
In particular, AttH first calculates the two predicted values for the tail entity by relation-specific hyperbolic reflecting and rotating the head entity. The two predictions are then combined into the final tail prediction through an attention mechanism with a relation-specific attention weight. The model is eventually optimised so that the true tail entity embedding is the closest to the prediction compared to the false ones.
In Atth, the mapping of hyperbolic spaces is able to better represent hierarchical relationships, thus Atth achieves good accuracy in relatively low dimensions.

\medskip
\noindent \textbf{Negative Sampling}
Negative sampling is a widely used technique in KGE that aims to improve the performance KGE models \cite{kamigaito2022comprehensive}. The key idea of negative sampling is to sample negative triples that do not exist in the KG and use them to train the KGE model along with positive triples. By doing so, the model learns to differentiate between positive and negative triples and improves its ability to predict missing relationships in the KG. The learning objective is usually set as maximising the difference between positive triple scores and negative triple scores, so that the positive triples are assigned higher scores and negative triples are assigned lower scores.
In this paper, we also explore the effectiveness of negative sampling in KGE by training with different negative size.

\medskip
\noindent \textbf{Multi-Layer Perceptron.}
A Multi-Layer Perceptron (MLP) classifier is a type of artificial neural network that is commonly used for classification tasks. The MLP consists of multiple layers of interconnected nodes or neurons that are organized into an input layer, one or more hidden layers, and an output layer. Each neuron in the network receives input from the neurons in the previous layer, and computes a weighted sum of those inputs using a set of learned weights. The weighted sum is then passed through an activation function, usually the sigmoid function or the ReLU function, to produce an output. The output of the final layer is used to make a classification decision. 
The weights of the MLP are learned through a process called backpropagation, which involves adjusting the weights to minimize a loss function that measures the difference between the predicted output of the network and the true output. MLP is used in many industrial applications such as abnormality detection, predictive quality maintenance~\cite{RAMANMR2020100393}.

\section{Method}
\label{sec:method}

\begin{figure*}[t]
\centering
\includegraphics[width=.95\textwidth]{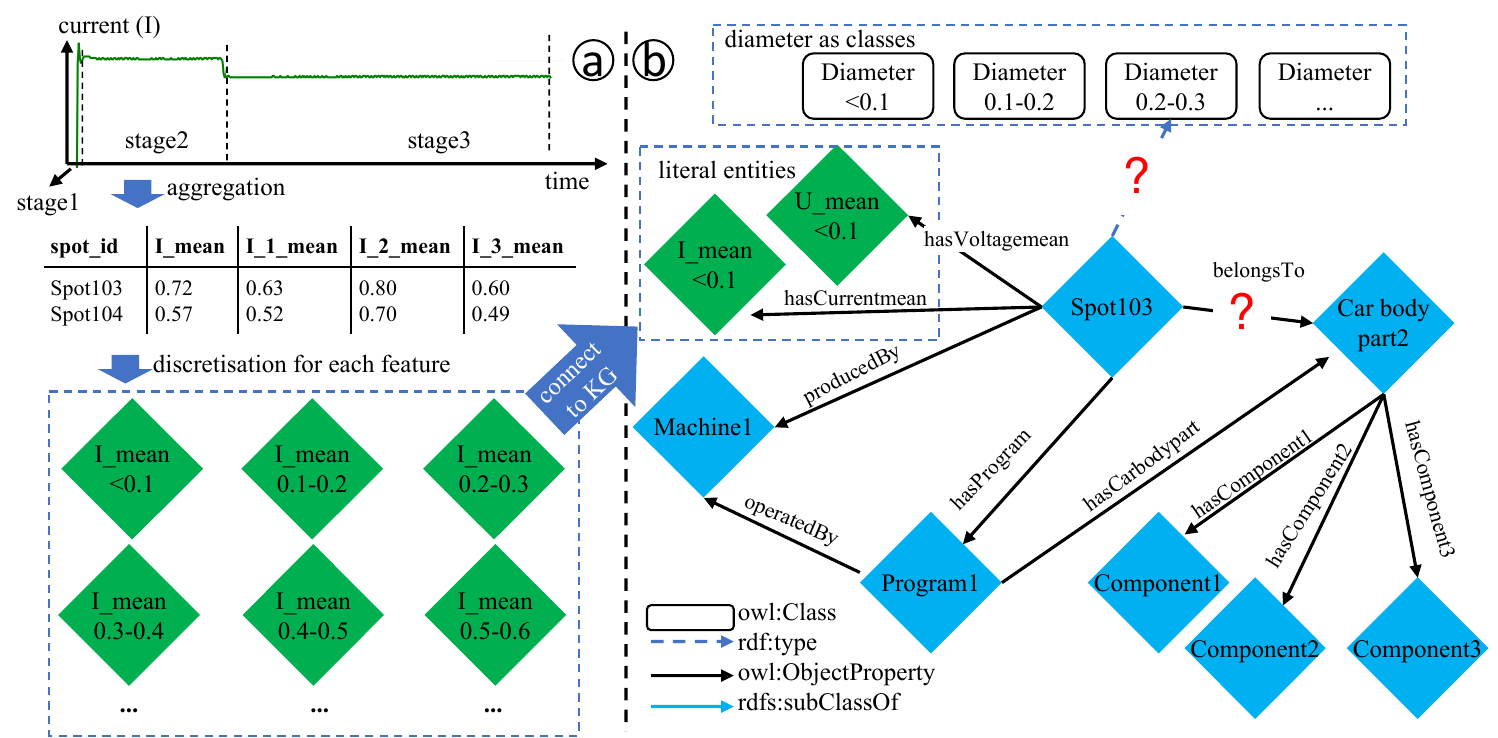}
\caption{\normalfont{(a) Procedure of literal embedding (b) Partial illustration of the welding KG}}
\label{fig:weldingKG}
\vspace{-2ex}
\end{figure*}

\noindent \textbf{Welding KG construction.} 
 A welding KG is constructed from the table data. We have used welding-related information, such as time of welding processes, welding machines, welding programs, and welding parameters (e.g., voltage, current, resistance). The constructions are conducted on welding spots and the car body and diameters. We transform the values of the welding data table into entities and the relationships between these entities as edges in the KG.
 Fig.\ref{fig:weldingKG}a shows the construction of literal entities, which are entities generated from numeric values. 
Based on the mean values, the current and voltages will be discretised by value ranges, such that all the numerics values are turned into entities. Discretised values such as machines and program id can be directly converted into entities.
 Fig.\ref{fig:weldingKG}b shows the Welding KGs, which are constructed from the data table with the form of the table~\ref{table:weldingdata}. And the two main research questions of our work.
Question 1 is the classification of the welding spot to the diameter classes. Question 2 is to predict the link between the carbodies and the welding spot, to find the correct carbody of the welding spots.

\begin{figure*}[t]
\centering
\includegraphics[width=.9\textwidth]{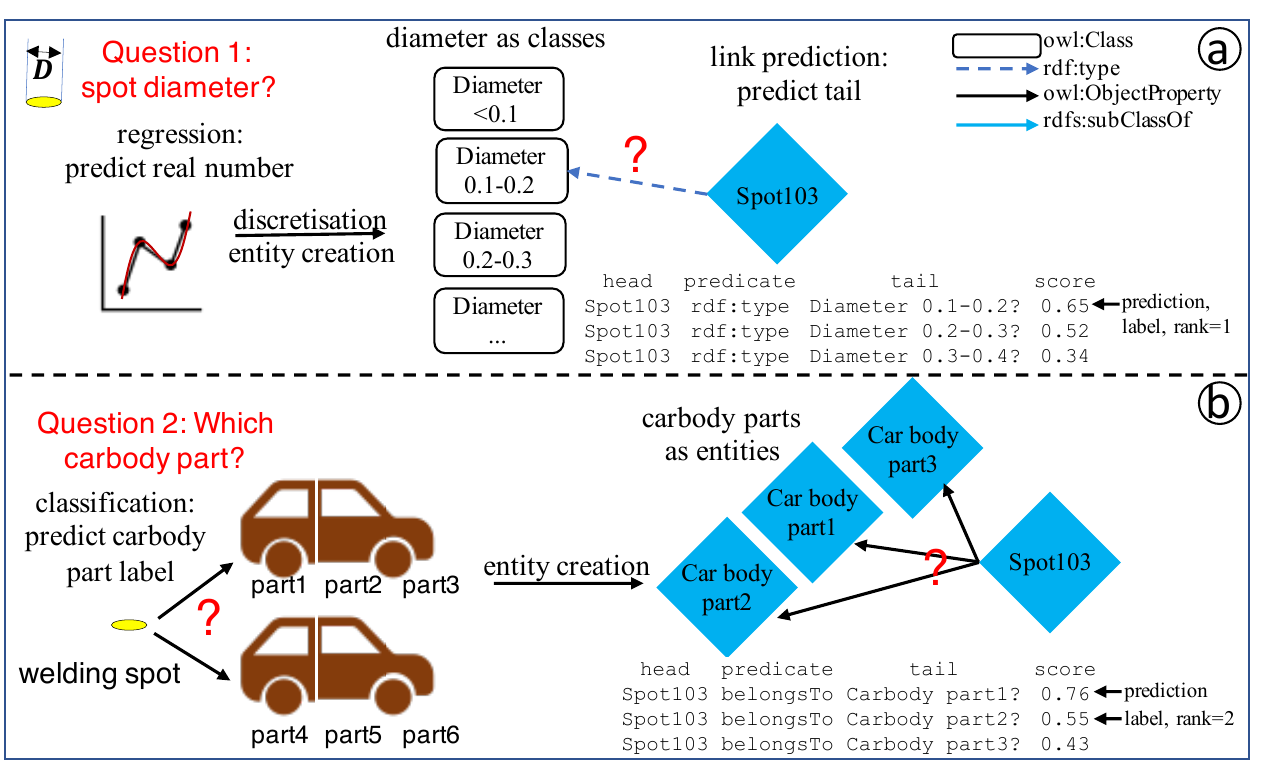}
 \vspace{-5mm}
\caption{Formulating regression and classification problems as link prediction problems.}
\label{fig:weldiproblemformulationngKG}
\vspace{-2ex}
\end{figure*}

\medskip \noindent \textbf{Problem formulation.}
Given the information of the welding spot, including the machine ID, program ID and their literal features such as voltage, current, resistance, welding time, welding power, the two research questions are to predict the carbody of the this welding spot and the diameter of the welding spot.
To make KGE applicable to the problem, we reformulate the two questions of the quality monitoring in the use case (Fig.~\ref{fig:weldiproblemformulationngKG}):
Q1: The Spot diameter prediction was a regression problem based on the welding data to predict the real values for the diameters size. Due to resolution when measuring the spot diameter, we discretise the diameters into different diameter classes and constructed the entities based on the diameter classes. We then predict the link between welding spots and the diameter classes. Considering the fixed differences between the neighbouring diameter classes, we use the mean differences between the diameter classes and calculated rmse based on the differences. 
Q2: For carbody classification, we conducted similar reformulation. Difference is that the carbodies are already discretised values, so we simply create the entities based on carbodies. Then the question is converted to predict the link between the carbody entities and the welding spots. With both of the reformulated research questions, we can apply KGE models, and their results become comparable to that of the original regression and classification problems.

\medskip
\noindent \textbf{KG construction from tabular data.}
The welding data are original in tabular form extracted from relational databases.
Since the tabular data are very extensive (over 200 columns) and contain many columns not closely related to the operation (e.g., unused machine settings), we need to construct welding KGs with relevant information describing welding operations. 
We construct KGs as the following steps: 
 (1) Remove all the empty columns and columns with only unique value, since these columns are not distinguishing information and are thus redundant for the welding knowledge graphs. 
(2) Choosing the most representative features based on domain knowledge from welding. Those representative features are to be put in the welding KG. 
(3) Process the literal features to be converted into KGs. Since many KGE approaches do not consider the literals when embedding the entities, we adapted literal-embedding approaches from previous works to convert the literals into entities.~\cite{10.1007/978-3-030-30793-6_20}
(4) Identify the entities and relationships: Look for the unique entries and their attributes in the table as the entities in the KGs. These relationships will be the edges in the KG and connect the entities. For example, one operation with id 1 was conducted on machine id 2, then the entities should be operation1 and machine2, with relationship ``conducted\_on\_machine''.

\medskip
\noindent \textbf{Literal handling.}
We did the following steps for literal embeddings inspired by~\cite{Wang2022AugmentingKG}:
aggregation, value discretisation, entity creation and linking. In the aggregation step, the sensor measured values are aggregated into the mean values of the three stages and the overall mean values in real numbers. Then in the discretisation step, we discretise the real values into different ranges. And then we create entities based on the discretised ranges. Then we link all the created literal entities with other entities.

\begin{figure*}[t]
\centering
\includegraphics[width=\textwidth]{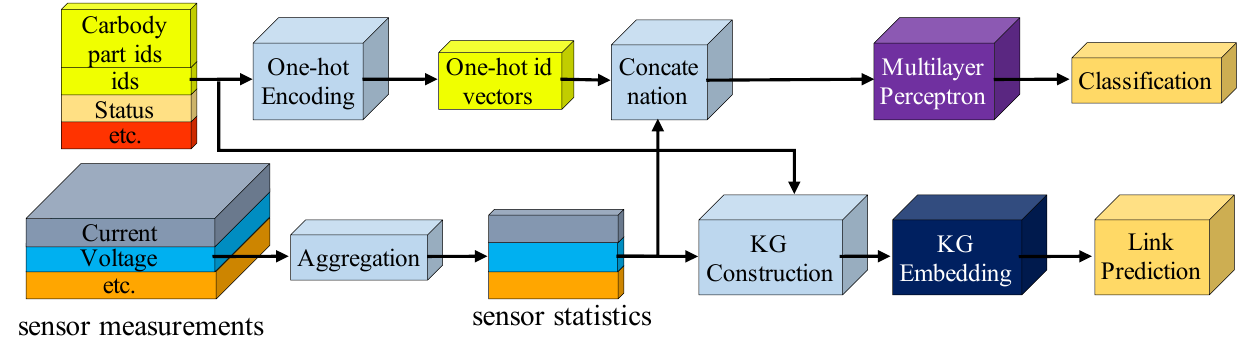}
\caption{\normalfont{The workflow of compared methods: the same raw data from relational tables go through different preprocessing (one-hot encoding or KG construction) and modelled by MLP or KGE methods, solving classification or link prediction problems for answering Q1 and Q2.}}
\label{fig:models}
\vspace{-2ex}
\end{figure*}

\medskip 
\noindent \textbf{Multi-layer Perceptron Classifier.}
In our work, MLP classifier is used to predict whether there exist connection between welding spot and the carbody or the diameter class, based on the information from welding knowledge graphs related to the welding spot and the carbody. Fig.\ref{fig:models}a shows the structure of the MLP classifier with embedding layer mapping each entity to a low-dimensional vector. 
The input of the MLP classifier is all the known parameters that belong to the welding spot, for example the welding machine, weling program, welding current, welding voltage etc. Their one-hot encoding will be fed to the MLP classifier. The output of the MLP classifier is the one-hot encoding of the carbodies or encoding of the diameters.

\noindent This model is the most basic model used in the quality monitoring and works as a baseline model for this use-case. Since the traditional manufacturing quality approach is totally different and not predictable, we can not compare directly with traditional method but rather compare all the machine learning approaches. \looseness=-1

\medskip 
\noindent \textbf{Knowledge Graph Embedding.}
As the development of knowledge graph embedding models in the recent years, there are various KGE models focusing on capture the information in the graph structural data.  Our KGE models are based on the famous models TransE which treat the entities and relations based on vector translation, DistMult model which treat relation with matrix factorization, AttH which maps the vector into hyperbolic space and calculate the score based on hyperbolic space vector. Those models shows good performance on open dataset and possess good generalization capability. Fig.~\ref{fig:models}b contains the KGE models architecture. The KGE model will embed all the entities and relations into a look-up table with a embedding layer. Afterwards the score or the distance of the input triple will be optimized based on the score functions tailored to different models. The input data of the model are triples in the form of (head, relation, tail) representing the single fact from the welding knowledge graphs. The output of the model is the score based on the distance of the triple, usually with the distance in the form of $d(h+r, t)$ where head entities embeddings are combined with relation embeddings in the model specific way, and then compare the vector distance with tail entity. The smaller the distance is, thus the higher is the score of the triple is. The training objective of KGE models is to maximize the scores of the input triples while minimize the scores of the non-existing triples by building the loss function as $Loss = s(h,r,t) - s(h,r,t*)$ where $t*$ represents all the negative sampling triples. To evaluate the KGE model after training, the ranks of correct triples will be calculated. For each correct triple in the welding knowledge graph, the tail will be replaced by other triples and the rank of correct tail will be calculated with respect to the other false tails. 

\medskip 
\noindent \textbf{Metrics and their meanings in the use case.} 
We consider 4 metrics for our questions: Acc(Hits@1), Hits@GroupBy3, \textit{nrmse}, MRR.

\noindent \textit{Acc(Hits@1)} represents the percentage of correct entities that are predicted correctly by assigning them the highest scores. In our case this can be understood as prediction accuracy and we use the Hits@1 to represent the accuracy. It is calculated by the percentage of entities having highest rank in the evaluation. In our use case, the Hits@1 represents the accuracy of the prediction. The range of the Hits@1 is between 0 and 1, where higher score represents the better performance.

\noindent \textit{Hits@GroupBy3} represents the percentage of correct entities that are predicted correctly when conducting a prediction in the group of 3 welding spots. In the testing, 3 carbody parts are grouped together and the prediction is considered correct if the predicted carbody part is in the same group as the ground truth carbody part. In our use case, the Hits@GroupBy3 is a less strict metrics than Hits@1, but it derives from the industrial scenario where the testing is conducted. It also represents the accuracy of the prediction.

\noindent \textit{nrmse} is the shortcut for normalised-root-mean-square-error, which was used to calculate the accuracy of diameter prediction. The metric is still applicable after we reformulating the problem into the link prediction problem, since the average diameter difference between the different diameter classes are known. It is calculated, as shown in the equation $ nrmse = \Sigma_i(D_i - \hat{D}_i)^2/\bar{D}$, as the square root of the mean value of the squared error.

\noindent \textit{MRR} is a measure of the average rank of the first correct entity or relation among all possible entities or relations. It is computed as the reciprocal of the rank of all the correct tail entities. MRR is more on the research purpose and used to compare the performances of different KGE models.

\section{Evaluation}
\label{sec:eva}

This section compares multilayer perceptron (MLP) with mainstream KGE methods and its KGE-MLP variants to investigate whether and to what extent KGE methods are useful for the two questions of interest in our use case. We also hope to shed light on KGE for industrial applications via discussions on the KGE performance and the KG characteristics.

\subsection{Experiment Settings}

\noindent \textbf{Baselines.}
We compare three mainstream KGE methods with the representative classic ML method MLP. The KGE methods include translational models in Eucledian space: TransE, RotatE, and hyperbolic model with attention mechanism: AttH. We select TransE and RotatE because they are representative methods, and are proven to have good generalisability to datasets and applications other than the settings under which they are developed~\cite{lin2015learning}. We select AttH as a representative method for more sophisticated KGE with attention mechanism and non-Euclidean embedding space.
We compare all KGE methods with MLP because MLP are widely applied in industrial applications, and have been proven to be universal approximators~\cite{pinkus1999approximation}.

\noindent In addition, we compare with a special type of KGE method proposed in~\cite{iswc2019inuse}, because this paper is a most close work on KGE for real-world use cases. The KGE method in this work is a combination of MLP and KGE method (thus is referred to as \textit{KGE-MLP}) to binary classification of triples (referred to has prediction triples),
MLP is used for learning the probability of the prediction triples as well as the embeddings of the head and tail entities in the prediction triples, and
KGE is used for learning all entity and relation embeddings. 

\medskip \noindent \textbf{Dataset details.}
We randomly select welding data in relational tables related to 2000 records of welding operations from a large database with over 260 k welding operations. We consider 2000 records is a balanced number to meaningfully test the method performance, while not spending excessive training time.
From over 200 features of the data, we removed certain features, such as constant values, NaN values, and other features according to domain knowledge.
The resulting dataset contains 31 meaningful features and 2000 rows, including ids for machines, programs, status, and aggregated values of sensor measurements.
Then the data are used for MLP after one-hot encoding, or used for KGE after KG construction, according to the procedures elaborated in Sect.~\ref{sec:method}.
The KG dataset contains 44801 triples, including 3342 entities, 26 relations, among which, 18 machines, 181 programIDs, 613 carbodies, 327 are literal entities.

\medskip \noindent
\textbf{Data splitting}. We split the table data following the standard 80\%/10\%10\% for train/validation/test, and do the following modifications for the KGE model. To ensure a fair comparison between MLP and no leakage appears, we firstly split the table data into train/validation/test dataset and then conduct the conversion from table data into KGs. 
The testset only contains triples with entities of \textit{welding spot}, \textit{carbodypart} and \textit{diameter}.
The negative triple samples are generated randomly following the negative sampling of existing KGE approaches.

\begin{table}[t]
    \centering
    \setlength{\tabcolsep}{2mm}
    \caption{Model performance comparison on answering Question 1 (Q1) and Question 2 (Q2). The best results are marked bold, and second best results marked by underlines. The experiments are repeated 5 times and results are reported as mean $\pm$ std. Note that MLP performance is not usable according to domain expert.}
    \vspace{-2mm}
    \resizebox{\textwidth}{!}{
    \begin{tabular}{lccccc}
    \toprule
 &            &  \textbf{MLP}   &  \textbf{TransE}   & \textbf{RotatE}   & \textbf{AttH}   \\ \hline
  &         Acc(Hits@1)  &  \underline{0.39} $\pm$ 0.01   &  \textbf{0.42} $\pm$ 0.02  &  0.25 $\pm$ 0.01  &  0.31 $\pm$ 0.05 \\
  &         MRR &  -   &  \textbf{0.65} $\pm$ 0.01  &  0.49 $\pm$ 0.00 &  \underline{0.57} $\pm$ 0.04  \\ 
Q1 &         \textit{nrmse}  & \textbf{0.05} $\pm$ 0.01 &  \underline{0.06} $\pm$ 0.00  &  0.08 $\pm$ 0.01 & \underline{0.06} $\pm$ 0.01  \\
 & time$_{train}$ & 120.6 $\pm$ 15.2 s & 660.1 $\pm$ 30.1 s & 1022.9 $\pm$ 80.5 s & 1829.1 $\pm$ 100.1 s \\ 
 & time$_{test}$ & $<$ 0.03s & $<$ 0.03s & $<$ 0.03s & $<$ 0.03s \\ \hline
 &         Acc(Hits@1) &  \underline{0.61} $\pm$ 0.01     &  \textbf{0.64} $\pm$ 0.01   & 0.52 $\pm$ 0.01  & 0.53 $\pm$ 0.03   \\ 
 &         MRR &  -    &  \textbf{0.77} $\pm$ 0.01   & 0.69 $\pm$ 0.01   & \underline{0.70} $\pm$ 0.01   \\  
Q2 &         Hits@Groupby3 &  -    &  \textbf{0.85} $\pm$ 0.01   & 0.81 $\pm$ 0.01 & \underline{0.79} $\pm$ 0.03   \\
 & time$_{train}$ & 117.1 $\pm$ 13.2 s & 822.6 s $\pm$ 101.7 & 1722.7 $\pm$ 152.7 s & 4266.9$\pm$ 234.5 s \\ 
 & time$_{test}$ & $<$ 0.03s & $<$ 0.03s & $<$ 0.03s & $<$ 0.03s \\ 
 \bottomrule
    \end{tabular}
    \label{table:performance}
    }
    \vspace{-4ex}
\end{table}

\medskip \noindent
\textbf{KGE setting.}
In this setting, negative samples are generated by corrupting existing triplets in the training set.
In the case of Q1, the tails in negative samples are  the wrong diameter classes.
In the case of Q2, the tails in negative samples are the wrong carbody part entities. 
The embedding size is experimented on a search space of  \{16, 32, 64, 128, 256\} and we choose the best hyper-parameters based on the MRR on the validation set.
We chose 128 or 256 as the best trade-off between efficiency and the accuracy. 

\medskip \noindent
\textbf{KGE-MLP setting.} 
The special variant of KGE-MLP models are tested with TransE \cite{transe}, DistMult \cite{yang2015embedding} and HolE \cite{HolE}, following~\cite{iswc2019inuse}.
The embedding size is experimented on a search space of  \{16, 32, 64, 128, 256\} and we chose 128 as the best trade-off between efficiency and the accuracy.

\subsection{Results and Discussion}

\noindent
\textbf{Results and Discussion on Question 1.}
From the results of MLP and KGE on Question 1 (table~\ref{table:performance}), we see that TransE performs the best in terms of Hits@1,  and MLP performs the best in terms of \textit{nrmse}. 

\medskip
\noindent
\textit{Acc(Hits@1).} 
For Q1, Hits@1 means the percentage of the correctly predicted diameter classes, therefore it represents the accuracy of diameter prediction, making it the most important metric in the use case.
In terms of Acc(Hits@1), TransE outperforms all other KGE and outperforms MLP by 6 \%. 
Important to note, for industrial applications, a prediction accuracy of only 0.42 is not sufficient, because industrial applications typically expect high accuracy. In our use case, we consider an accuracy at least over 80\% can make the solution usable, and over 90\% makes the solution good. 
However, we should not be too pessimistic, and can relax the evaluation metric by resorting to \textit{nrmse}.
The MLP model performs not well on Hits@1 for the diameter class prediction. As only 0.39 for the prediction accuracy, meaning not even half of the diameter is predicted correctly. Among all KGE models, \textit{TransE} achieves the best results and is also better than the MLP model results, with 0.42 accuracy. Other two KGE models RotatE and AttH didn't show improvements compared with the MLP model, but they still manage to predict around 1/4 of the diameter classes correctly.

\medskip
\noindent
\textit{MRR.} For common KGE problems, MRR is also a important performance indicator, and mostly correlates with Hits@1. In the use case, the MRR indicates that TransE is the best KGE model with an MRR of 0.65 and AttH is the second best with 0.57. There exist no MRR for MLP and thus they are not comparable in terms of MRR.

\medskip
\noindent
\textit{nrmse.} 
Here the \textit{nrmse} is calculated by converting the diameter classes back to real values then using the equation $\Sigma_i(D_i - \hat{D}_i)^2/\bar{D}$.
It indicates the  average prediction error normalised by the mean value of the prediction target and can be understood as relative percentage error. 
We see that in terms of \textit{nrmse}, MLP model is the best one, while TransE and AttH are comparable to MLP, and their prediction errors are 5\% to 6\%.
We postulate the reason that diameter prediction was originally a regression problem, and classic ML method such as MLP is well-suited for regression problems. 

\noindent Considering the industrial adoption, we think the relative percentage error of 5\% - 6\% are both acceptable for industrial adoptions in our use case and thus both MLP and KGE can be adopted in principle.

\medskip
\noindent
\textit{Time.}
In terms of training time, MLP consume much less than than all KGE models, while TransE consumes the least time among the KGE models. 
In terms of test time, all models are comparable.
Considering industrial adoption, in the case where training time is critical, MLP has an advantage. In mose cases, models in industrial applications are pre-trained and the test time is more important. In this regards, all models do not have adoption issues in terms of test time.

\medskip
\noindent
\textbf{Results and Discussion on Question 2.}
From the results of MLP and KGE models on Q2 (Question 2), we can see that TransE also performs the best in terms of Acc(Hits@1), Hits@3Group and MRR (lower part of Table~\ref{table:performance}).

\medskip
\noindent
\textit{Acc(Hits@1).} 
This metric means the percentage of the correctly predicted carbody parts. In terms of Acc(Hits@1), TransE is the best model and outperforms MLP model by 5\% (relative). 
We see that for the carbody part problem, which can be regarded as classification problem, TransE is better suited.
However, a prediction accuracy of 0.61 means 61\% carbody is corrected predicted, which is still not sufficient for industrial application. We also consider relax the prediction and rely on Hits@Groupby3.

\medskip
\noindent
\textit{MRR.} 
Similar to Question 1, MRR is highly correlated with Hits@1 and TransE is the best KGE model.

\medskip
\noindent
\textit{Hits@Groupby3.} Hits@Groupby3 is  a metric we propose that is similar to Hits@3. It is  calculated by first splitting the carbody parts into groups where each group has 3 carbody parts then count whether the predicted carbody is within the group.
Hits@Groupby3 relaxes the prediction by requiring to predict the correct carbody part group (with 3 carbody parts) instead of one carbody.
In industrial practice we can also rely on Hits@Groupby3, because we can consider the carbody part groups as the minimal unit of quality monitoring and and evaluate the quality by the groups.
With this new metric Hits@Groupby3, we can see all the KGE models have relative good performance, about 0.80. TransE is still the best model with 0.85.

\medskip
\noindent
\textit{Time.}
In terms of training time, the models consume time differently, with MLP consuming much less than than all KGE models, while TransE consumes the least time among the KGE models. 
In terms of test time, all models are comparable in the Q2 prediction.
Considering industrial adoption, in the case where training time is critical, MLP has an advantage. In the case where the models only conduct inference, the TransE with higher performance si advantageous. In this regards, all models do not have adoption issues in terms of test time.

\begin{wraptable}{r}{.6\textwidth}
\vspace{-7ex}
\caption{Comparison with KGE-MLP Models. The KGE-MLP variants are marked with*.}
\label{tab:kge+mlpresults}
\setlength{\tabcolsep}{1mm}
\resizebox{.6\textwidth}{!}{
\begin{tabular}{cccccc}
\toprule  
& Metric & TransE & TransE* & DistMult* & HolE* \\ \hline
 & Acc(Hits@1) & \textbf{0.42} & 0.17 & \underline{0.22} & 0.21 \\
Q1 & MRR & \textbf{0.65} & 0.45 & \underline{0.48} & \underline{0.48} \\ 
 & \textit{nrmse} & \textbf{0.06} & 0.11 & \underline{0.09} & 0.10 \\ 
 \hline
& Hits@1 & \textbf{0.64} & 0.34 & 0.34 & \underline{0.37} \\
Q2 & MRR & \textbf{0.77} & 0.48 & \underline{0.52} & 0.41 \\
& Hits@GroupBy3 & \textbf{0.85} & 0.45  & 0.46 & \underline{0.52} \\
\bottomrule
\end{tabular}
}
\vspace{-3ex}
\end{wraptable}

\medskip
\noindent
\textbf{Comparison with KGE-MLP models}.
We observe from Table~\ref{tab:kge+mlpresults} that all the KGE-MLP variants performs worse than their KGE counterparts in terms of Acc(Hits@1), MRR, and \textit{nrmse}.
With 0.34 for accuracy and 0.52 for MRR for carbody prediction and 0.22 for accuracy and 0.48 for MRR for diameter prediction, the DistMult based MLP model shows no performance improvement over the baseline MLP model regarding the carbody prediction. Possible reason could be the MLP  in this model can not capture the welding information well.
The special KGE variants were suitable for the problem of~\cite{iswc2019inuse},  but they seem to be not well-suited for the questions in our case. 

This results indicate that the welding KG and KGE based MLP can capture the information of the welding data.
However, due to the model design it may not work as well as the KGE models.

\begin{wraptable}{r}{.6\textwidth}
\vspace{-6ex}
\caption{Ablation study on literals, models without literals are marked with $\dagger$}
\label{tab:ablationstudy}
\setlength{\tabcolsep}{2mm}
\resizebox{.6\textwidth}{!}{
\begin{tabular}{cccccc}
    \toprule
        &  & TransE & MLP & TransE$\dagger$ & MLP$\dagger$ \\ 
        \hline
        & Acc(Hits@1) & \underline{0.42} & 0.39 & \textbf{0.45} & 0.36  \\ 
        Q1 & \textit{nrmse} & 0.06 & 0.05 & 0.04 & 0.05  \\  
        & MRR & \underline{0.64} & - & \textbf{0.68} & -  \\ \hline
        & Hits@1 & 0.64 & 0.61 & 0.53 & 0.49  \\ 
        Q2   & MRR & \textbf{0.77} & - & 0.70 & -  \\          
        & Hits@GroupBy3 & \textbf{0.85} & - & 0.78 & -  \\
        \bottomrule
    \end{tabular}
}
\vspace{-3ex}
\end{wraptable}
\medskip \noindent 
\textbf{Ablation study on literals.}
Table~\ref{tab:ablationstudy} shows the results of the ablation study regarding the literals. We can see that for answering Q1, performance of TransE is even worse than TransE$\dagger$ (without literals), while MLP performs better with literals. We repeated the experiments 5 times and this result is persistent. We postulate the reason is that the \textit{Diameter} class are isolated in the KG, as shown in Fig.~\ref{fig:weldingKG} that the {\texttt{rdf:type}} is the only link connecting \textit{Diameter} with the rest of the KG.
This makes  it is difficult to learn the correct links between the \textit{Diameter} and \textit{Spot}.
For answering Q2, we see both  MLP  and TransE models show some performance degradation without literals. 
This is expected. We can see from Fig.~\ref{fig:weldingKG} the \textit{Carbodypart} is ``well'' connected to the rest of the KG via many other links, and thus it does not suffer the same issue as in Q1. 

\medskip \noindent 
\textbf{Recommendation for industrial adoption.}
For answering Q1, MLP has slightly better performance than TransE in terms of \textit{nrmse}, which is the most meaningful metric for industrial adoption, but the different of 5\% and 6\% error is marginal. Considering MLP has less training time and it is easier to understand for domain experts than KGE, it is still preferred than TransE.
For answering Q2, TransE has better results than MLP. Although none of the methods are directly applicable considering  the prediction accuracy (Hits@1), we can relax the condition by introducing Hits@Groupby3, which requires to group the carbody parts first by 3 and judging the quality for each carbody part group instead of for each carbody. This is a good news for industrial adoption since the carbody classification problem is challenging for classic ML due to the high number of label classes.
Overall, we consider the adoptabilty of KGE in industrial applications is promising, although not perfect. The adoptability is increased by the adaptation of problem formulation, handling numerical literals and introducing new metrics.
\section{Discussion on General Impact and Related Work}
\label{sec:uptake}

\noindent \textbf{Related work of KGE.}
Representative KGE models include translational models in Eucledian space, such as
TransE~\cite{transe}, RotatE~\cite{sunrotate}, and  model in hyperbolic space with attention mechanism: AttH~\cite{chami2020low} as the KGE models for quality monitoring. 

\medskip
\noindent \textbf{Related work of KGE applications.}
In the past work, there exist some applications of KGE in real-world applications. Such as the application in  natural language processing \cite{Santini2022}, ecooxicological effect prediction \cite{iswc2019inuse}, application in biological area~\cite{mohamed2021biological}. 
There exist rather less work on KGE in industrial applications.
Our work is an attempt to test whether and to what extent KGE can be used in industries, especially traditional industries such as manufacturing, rather than internet industries.

\medskip \noindent \textbf{Usability by Stakeholders.}
Our work is grouped into the Bosch research for data-driven solutions for manufacturing condition monitoring, and under the umbrella of new generation manufacturing monitoring solutions based on neuro-symbolic methods. The project spans over three sub-projects: the resistance spot welding quality monitoring, process optimisation for hot-staking, and plastic data analytics. The project collaborates with several factories in Germany, aiming at smart equipment analytics, process optimisation, etc.
Currently the solution is under evaluation environment of both Bosch research and factories, and received positive feedback, where the stakeholders are data scientists, semantic experts, R\&D engineers, etc. After the evaluation and prototyping, the solution will be moved to the factories. 

\medskip \noindent \textbf{General Uptake.}
We provide our \textit{scripts} of ML and KGE and \textit{anonymised welding KG dataset} in the open GitHub repository.
Our scripts of ML and KGE should provide important resource for reproducing the methods and results.
Our goal is to facilitate neuro-symbolic research that combines semantic technologies and ML for industrial applications, especially for manufacturing industry.
We hope our provided resource can inspire research in the community of neuro-symbolic reasoning, semantic technologies, graph embedding, etc. and advance the state of art in these domains. We observe that the most open-source KG datasets are about common sense domains, and few are about industries such as manufacturing.
We thus hope that our dataset can help more researchers to connect their research on KG to industrial cases.

\medskip \noindent \textbf{Scalability and benefits.}
We have tested our solutions and recorded the consumed resource and time.
The results (Table~\ref{table:performance}) show that the test time which is critical for scalabilty in industry is acceptable.
The benefits of this work are quite obvious. Data-driven solutions help to reduce cost and waste and increase quality monitoring covering, by reducing or eventually replacing the conventional destructive methods that destroy samples of welded car bodies. Further benefits are that the work is an attempt of neuro-symbolic reasoning for manufacturing that hopes to inspire more research directions.

\medskip \noindent \textbf{Risks and opportunities.}
The risks here are that the prediction of both diameters and carbody parts are not perfect. This needs to be documented in the quality monitoring system if the technology is equipped in such a system. Despite this, the risks are comparable to conventional methods and other data-driven solutions, and are manageable by correct understanding of the prediction results and adopting the measures such as safety coefficients.
The research provide opportunities for further investigation and improvement of the solutions on other manufacturing condition monitoring questions, datasets, and use cases.
This research also provide opportunities for researchers working on industry digitisation and the AI application.

\section{Conclusion and Outlook}
\label{sec:conclusion}

\noindent \textbf{Conclusion and outlook.}
This paper investigats to whether and to what extend KGE can be applied for Bosch welding quality monitoring. We compared KGE methods with MLP on two important questions in manufacturing quality monitoring.
To make KGE applicable for our industrial questions, we adapted the KGE methods in these aspects: we formulated classic ML problems of classification and regression to link prediction, proposed strategies for handling literals, including sensor measurements and diamters, and
introduced the performance metric \textit{rmse} and Hits@GroupBy3.
The KGE are not directly applicable if we only consider the original metrics of KGE such as Hits@1 and MRR, but after relaxing the prediction task by adopting \textit{rmse} and Hits@GroupBy3, the adoptability of KGE is increased and we give recommendations on the adoption.

This paper is under the umbrella of Neuro-Symbolic AI for Industry 4.0 at Bosch~\cite{DBLP:conf/semweb/ZhouTZZHZYT0GK22} that aims at enhancing manufacturing with both symbolic AI (such as semantic technologies~\cite{yahya2023semantic}) for improving transparency~\cite{zheng2022executable}, and ML for prediction power~\cite{klironomos2023exekglib}.
As future work, we plan to investigate larger datasets and improve the KGE performance. We work closely with colleagues from factories and will investigate further the adoptability of the technology.

\medskip
\noindent\textbf{Acknowledgements}
The work was partially supported by EU projects Dome 4.0 (953163), OntoCommons (958371), DataCloud (101016835), Graph Massivizer (101093202) and enRichMyData (101093202) and the SIRIUS Centre (237898) funded by Norwegian Research Council .


%
\bibliographystyle{elsarticle-num}
\bibliography{iswckgewelding}
\end{document}